# Axiomatic Foundations for a Class of Generalized Expected Utility: Algebraic Expected Utility

**Paul Weng**
Paris 6 University, LIP6
8 rue du Capitaine Scott
75015 Paris, France

## Abstract

In this paper, we provide two axiomatizations of algebraic expected utility, which is a particular generalized expected utility, in a von Neumann-Morgenstern setting, i.e. uncertainty representation is supposed to be given and here to be described by a plausibility measure valued on a semiring, which could be partially ordered. We show that axioms identical to those for expected utility entail that preferences are represented by an algebraic expected utility. This algebraic approach allows many previous propositions (expected utility, binary possibilistic utility,...) to be unified in a same general framework and proves that the obtained utility enjoys the same nice features as expected utility: linearity, dynamic consistency, autoduality of the underlying uncertainty representation, autoduality of the decision criterion and possibility of modeling decision maker's attitude toward uncertainty.

## 1 INTRODUCTION

Axiomatic justification of a decision model is essential as it underlines the properties of the decision model and reveals what decision behavior it can or cannot describe. The seminal work presented in von Neumann and Morgenstern (1944) made expected utility (EU) the first axiomatized decision model. This axiomatization was then reformulated and simplified (Fishburn (1970); Machina (1988)). In von Neumann-Morgenstern (vNM) like axiomatizations, (objective) probabilities are assumed to be given. This assumption is relaxed in the axiomatization proposed by Savage (1954). In such a setting, (subjective) probabilities are constructed from the preferences of the decision maker if and only if some required axioms are satisfied. However the assumption that the uncertainty is represented by objective probability (vNM setting) or the requirement of axioms leading to the construction of subjective probability (Savagean setting) can be debatable in some situations, especially those where data are imprecise and/or scarce (Ellsberg (1961)). Moreover, even when the decision problem imposes probability as the natural uncertainty measure, it can be difficult to assess.

To tackle these difficulties, other models for uncertainty have been proposed as generalizations of probability: belief functions (Shafer (1976)), $\kappa$-rankings (Spohn (1988); Goldszmidt and Pearl (1992)), possibility theory (Dubois and Prade (1990)), symbolic probability (Darwiche and Ginsberg (1992)) to name a few. They are all instances of plausibility measures (not to be confused with plausibility functions) introduced by Friedman and Halpern (1995).

For all these uncertainty representations, decision models have been proposed (Jaffray and Wakker (1993); Giang and Shenoy (2000); Dubois and Prade (1995),...). Some of these already proposed criteria share some formal similarities. This fact has been recently underlined in Giang and Sandilya (2004) who axiomatically justified a general decision model based on symbolic probabilities. Earlier, Dubois et al. (1996) studied a general class of real-valued decision criteria based on triangular norms (Klement et al. (2000)). One of the motivations of our work is to formally highlight the similarities of this class of criteria exploiting generalized expected utility introduced by Chu and Halpern (2003a,b). This is a very general decision criterion, built on top of plausibility measures.

More specifically, we provide two axiomatizations in a vNM setting for a class of generalized expected utility: algebraic expected utility (AEU). Uncertainty representation is supposed to be given and modeled by plausibility measures valued on a semiring, which could be partially ordered. Approaches relying on semirings have proved to be able to offer a unified treatment to many variations of the same problem (see Gondran et al. (1984) for graph problems, Bistarelli et al. (1999) for constraint satisfaction problems and Perny et al. (2005) for Markov decision processes). The first axiomatization, similar to that of Luce and Raiffa (1957), is a reformulation in a vNM setting of the one proposed in Giang and Sandilya (2004). The sec-ond one inspired by that of Machina (1988) is based on simpler and more natural postulates but requires two solvability assumptions. This algebraic treatment allows the counterpart of expected utility to be formulated for many uncertainty representations, thus unifying many different models in a same general formalization. Moreover this unification emphasizes that these decision models all enjoy the same nice formal properties as expected utility: linearity, dynamic consistency allowing an easy integration of AEU in sequential decision making, especially in algebraic Markov decision processes (Perny et al. (2005)), autoduality of the underlying uncertainty representation, autoduality of the decision criterion and possibility of modeling decision maker's attitude toward uncertainty.

The rest of the paper is divided as follows. In Section 2 we introduce the required definitions: formalization of a decision problem, plausibility measures, semiring, binary scale and generalized expected utility. In Section 3 we present the two sets of axioms and the two representation theorems for algebraic expected utility. Then we underline its properties in Section 4. In Section 5 we compare our approach with related works. Finally we wrap up in Section 6.

## 2 FRAMEWORK

In decision making under uncertainty, an agent has to make a choice among a set of alternatives or acts that are functions from a set of states of the world to a set of consequences. A state of the world can be thought of as a complete description of the environment. Uncertainty about the exact consequence of an act results from the fact that the decision maker does not know in which state he/she is. It is thus modeled as uncertainty about the actual state of the world. This can be formalized by:

- $X = \{x_1, \ldots, x_n\}$ a finite set of consequences with a best consequence $\overline{x}$ and a worst one $\underline{x}$.
- $S$ a set of states of the world.
- $\mathcal{A}$ a set of acts defined as $X^S$.

The decision maker's preference relation (which is simply a binary relation) over the set of acts is denoted by $\succsim$, which reads "at least as good as". Acts will be denoted $f, g, \ldots$ Relation $\succsim$ restricted to constant acts (consequences) is denoted $\succsim_X$. Relations $\succ$ and $\sim$ denote respectively the asymmetric and the symmetric part of $\succsim$.

Events in this context are defined as elements of an algebra $\mathcal{F}$ of subsets of $S$, i.e. a set of subsets closed under finite union and complementation. Events will be denoted $A, B, \ldots$ We assume that uncertainty about the actual state is captured by plausibility measures (Friedman and Halpern (1995)). A plausibility measure Pl is a function from $\mathcal{F}$ to a (possibly partially) ordered scale $P$ such that $\text{Pl}(\emptyset) = 0_P$, $\text{Pl}(S) = 1_P$ and $\forall A, B \in \mathcal{F}, B \subseteq A \subseteq S \Rightarrow \text{Pl}(A) \geq \text{Pl}(B)$. Scale $P$ is assumed to be endowed with two internal operators $\oplus$ and $\otimes$ (the analogs of $+$ and $\times$ in probability theory), a (possibly partial) order relation $\geq$, and two special elements $0_P$ and $1_P$. Elements of this scale will be denoted $\lambda, \mu, \ldots$ We assume here that Pl is decomposable, i.e. $\text{Pl}(A \cup B) = \text{Pl}(A) \oplus \text{Pl}(B)$ for any pair $A, B$ of disjoint events and $\text{Pl}(A \cap B) = \text{Pl}(A) \otimes \text{Pl}(B)$ for any pair $A, B$ of plausibilistically independent events.

From now on, structure $(P, \oplus, \otimes, 0_P, 1_P)$ is assumed to be a semiring (see Gondran et al. (1984)).

**Definition 1** *A semiring* $(P, \oplus, \otimes, \mathbf{0}, \mathbf{1})$ *is a set $P$ with two binary operations $\oplus$ and $\otimes$, such that:*
$A_1$ $(P, \oplus, \mathbf{0})$ *is a commutative semigroup with $\mathbf{0}$ as neutral element $(a \oplus b = b \oplus a, (a \oplus b) \oplus c = a \oplus (b \oplus c), a \oplus \mathbf{0} = a)$.*
$A_2$ $(P, \otimes, \mathbf{1})$ *is a semigroup with $\mathbf{1}$ as neutral element, and for which $\mathbf{0}$ is an absorbing element (i.e. $(a \otimes b) \otimes c = a \otimes (b \otimes c), a \otimes \mathbf{1} = \mathbf{1} \otimes a = a, a \otimes \mathbf{0} = \mathbf{0} \otimes a = \mathbf{0}$).*
$A_3$ $\otimes$ *is distributive with respect to $\oplus$ (i.e. $(a \oplus b) \otimes c = (a \otimes c) \oplus (b \otimes c), a \otimes (b \oplus c) = (a \otimes b) \oplus (a \otimes c)$).*

In this setting, order $\geq$ is related to $\oplus$ and we have $\forall \lambda, \mu \in P, \lambda \geq \mu \Rightarrow \exists \mu' \in P, \lambda = \mu \oplus \mu'$. These requirements or some variants of them were also imposed in Darwiche and Ginsberg (1992); Friedman and Halpern (1995); Weydert (1994) from general scales and in Weber (1984); Dubois et al. (1996) for real scales. All these conditions are quite natural and are verified by many plausibility measures. Here are some examples. For each of them, we explicit the associated semiring. Probability is defined on semiring $(\mathbb{R}^+, +, \times, 0, 1)$. Qualitative possibility is defined on $(L, \max, \min, 0_L, 1_L)$ where $L$ is a finite qualitative scale with lowest element denoted $0_L$ and greatest element $1_L$. Quantitative possibility is defined on $([0, 1], \max, \times, 0, 1)$. $\kappa$-rankings is defined on $(\mathbb{N} \cup \{+\infty\}, \min, +, +\infty, 0)$. Symbolic probability is defined on $(\mathcal{S}, \boxplus, \boxtimes, \bot, \top)$ where $\mathcal{S}$ is a support set containing two special elements $\bot$ and $\top$ and operators $\boxplus$ and $\boxtimes$ play similar roles as $\oplus$ and $\otimes$.

From $P$, a particular binary scale $P_2$ can be constructed.
$$P_2 = \{\langle \lambda, \mu \rangle \in P \times P : \lambda \oplus \mu = 1_P\}.$$
Elements of $P_2$ will be denoted $\alpha, \beta, \ldots$ where $\alpha = (\alpha_1, \alpha_2), \beta = (\beta_1, \beta_2)$ with $\alpha_1, \alpha_2, \beta_1, \beta_2$ in $P$. The couples in this set can be interpreted as plausibility values. The first (resp. second) element of a couple would measure the plausibility of getting the best (resp. worst) outcome. Then a natural order $\geq_2$ can be defined over $P_2$ from $\geq$ of $P$, i.e.
$$\langle \lambda, \mu \rangle \geq_2 \langle \lambda', \mu' \rangle \Leftrightarrow (\lambda \geq \lambda' \text{ and } \mu' \geq \mu).$$
Operator $\oplus$ is extended as an operator on $P_2 \times P_2$ as follows: $\langle \lambda, \mu \rangle \oplus \langle \lambda', \mu' \rangle = \langle \lambda \oplus \lambda', \mu \oplus \mu' \rangle$. And operator $\otimes$ is extended as an operator on $P \times P_2$ as follows: $\lambda' \otimes \langle \lambda, \mu \rangle = \langle \lambda' \otimes \lambda, \lambda' \otimes \mu \rangle$. The overloading of the operators should not pose a problem as context will make clear if operands are in scale $P$ or scale $P_2$.

We now define the artifacts by which acts are compared. A simple lottery $\pi$, which is just a plausibility distribution,

is defined as a function from the set of consequences $X$ to the qualitative scale $P$, such that $\bigoplus_{x \in X} \pi(x) = 1_P$. The set of all such lotteries is denoted by $\Pi(X) = \{\pi \in P^X : \bigoplus_{x \in X} \pi(x) = 1_P\}$. Recursively we define the set of compound lotteries that are lotteries over lotteries: $\Pi^1(X) = \Pi(X)$ and $\Pi^k(X) = \Pi^{k-1}(\Pi(X)), \forall k > 1$. The set of all (simple and compound) lotteries is then denoted by $\Pi^\infty(X) = \cup_{k=1}^\infty \Pi^k(X)$. For ease of notation, $x$ denotes both an element of $X$ and the degenerated lottery $\pi_x \in \Pi(X)$ such that $\pi_x(x) = 1_P$ and $\pi_x(z) = 0_P, \forall z \neq x$. As the set of consequences is finite, a simple lottery $\pi$ can be written: $[\lambda_1/x_1, \ldots, \lambda_n/x_n]$ where $\bigoplus_{i=1}^n \lambda_i = 1_P$. And more generally, a compound lottery with $k$ outcomes is written: $[\lambda_1/\pi_1, \ldots, \lambda_k/\pi_k]$ where $\bigoplus_{i=1}^k \lambda_i = 1_P$ and $\forall i = 1 \ldots k, \pi_i \in \Pi^\infty(X)$.

Any act is associated with a unique lottery. For an act $f \in \mathcal{A}$, the associated lottery $\pi^f$ is defined by $\pi^f(x) = \mathrm{Pl}(\{s \in S : f(s) = x\}), \forall x \in X$. We assume that comparing acts is equivalent to comparing their associated lotteries. Then the preference relation over acts induces a preference relation over lotteries, which we also denote $\succsim$. From now on, we work directly on lotteries.

Using plausibility measures, Chu and Halpern (2003a,b) proposed generalized expected utility (GEU) as a decision criterion. In this model, a set $U$ endowed with $\succsim_U$ describes the tastes of the decision maker about consequences and a set $V$ endowed with $\succsim_V$, which is the valuation set for GEU, describes the preferences of the decision maker about acts. Then GEU is defined over an expectation domain $(U, P, V, \oplus^g, \otimes^g)$ where $\oplus^g : V \times V \to V$ and $\otimes^g : P \times U \to V$ are the counterparts of $+$ and $\times$ in probabilistic expected utility, and the four following conditions are required:
$B_1$ $(x \oplus^g y) \oplus^g z = x \oplus^g (y \oplus^g z)$,
$B_2$ $x \oplus^g y = y \oplus^g x$,
$B_3$ $1_P \otimes^g x = x$,
$B_4$ $(U, \succsim_U)$ is a substructure of $(V, \succsim_V)$.

For any act, GEU writes:
$$GEU(\pi) = \bigoplus_{x \in X}^g \pi(x) \otimes^g u(x)$$
where $u : X \to U$ is a basic utility assignment.

GEU is a very general decision model and as shown in their papers, GEU can represent any preference relation. As an example, the expectation domain of EU is $([0,1], [0,1], [0,1], +, \times)$. Exploiting this framework, we want to construct the counterpart of expected utility for plausibility measures satisfying requirements $A_1$-$A_3$.

## 3 ALGEBRAIC EU

Algebraic expected utility (AEU) is a GEU defined on the expectation domain $(P_2, P, P_2, \oplus, \otimes)$. It writes:

$$AEU(\pi) = \bigoplus_{x \in X} \pi(x) \otimes u(x)$$
where $u : X \to P_2$ is a basic utility assignment.

As $(P, \oplus, \otimes, 0_P, 1_P)$ is a semiring, conditions $B_1$-$B_4$ are naturally satisfied. Our framework makes clear how the expectation domain is constructed.

We now introduce the first set of axioms, which is a reformulation in a vNM setting of axioms presented in Giang and Sandilya (2004) and are similar to those of Luce and Raiffa (1957):

$R$ (Reduction of lotteries)
$$\forall x \in X, [\lambda_1/\pi_1, \ldots, \lambda_m/\pi_m](x) = \bigoplus_{i=1}^m \lambda_i \otimes \pi_i(x).$$

$C_1$ (Preorder) $\succsim$ is reflexive and transitive.

$C_2$ (Order over binary lotteries)
$\alpha \geq_2 \beta \Leftrightarrow [\alpha_1/\overline{x}, \alpha_2/\underline{x}] \succsim [\beta_1/\overline{x}, \beta_2/\underline{x}]$.

$C_3$ (Substitutability)
$\forall i = 1 \ldots k, \pi_i \sim \pi'_i, \lambda_i \in P$, s.t. $\bigoplus_{i=1}^k \lambda_i = 1_P \Rightarrow [\lambda_1/\pi_1, \ldots, \lambda_k/\pi_k] \sim [\lambda_1/\pi'_1, \ldots, \lambda_k/\pi'_k]$.

$C_4$ (Continuity on consequences)
$\forall x \in X, \exists \alpha \in P_2$ s.t. $x \sim [\alpha_1/\overline{x}, \alpha_2/\underline{x}]$.

Axiom $R$ allows compound lotteries to be reduced to simple lotteries. Under this condition, we have $\Pi(X) = \Pi^\infty(X)$. Axiom $C_1$ is a typical ordering assumption. The preorder is not assumed to be complete. Axiom $C_2$ enforces an order over binary lotteries, i.e. lotteries having only two outcomes, the best and the worst consequences. This can be considered quite natural as it says that lotteries with a high plausibility of getting the best consequence and a low plausibility of getting the worst one are preferred. Axiom $C_3$ states that replacing sublotteries by equivalent ones in a compound lottery yields equivalent compound lotteries. Axiom $C_4$ says that any sure consequence is equivalent to a binary lottery.

Based on this set of axioms, the following representation theorem for AEU can be stated:

**Theorem 1** *Under condition $R$, preference relation $\succsim$ over $\Pi^\infty(X)$ satisfies axioms $C_1$-$C_4$ iff there exists a basic utility assignment $u : X \to P_2$ such that*
$\forall \pi, \pi' \in \Pi^\infty(X), \pi \succsim \pi' \Leftrightarrow AEU(\pi) \geq_2 AEU(\pi')$.

**Proof.** We only give a sketch of the proof, which is quite classic. The necessary part follows from the definition of AEU and binary scale $P_2$. For the sufficient part, axiom $C_4$ says that for any degenerated lottery, i.e. any consequence, one can find an equivalent binary lottery. Under condition $R$, axiom $C_3$ entails that any lottery can also be reduced to a binary lottery. Axiom $C_1$ states that comparing lotteries is equivalent to comparing their associated binary lotteries. Finally axiom $C_2$ gives the order over binary lotteries. ∎

We now present the second set of axioms.

$D_1$ (Preorder) $\succsim$ is reflexive and transitive.

$D_2$ (Non triviality)
$[\lambda/\overline{x}, \mu/\underline{x}] \sim [\lambda'/\overline{x}, \mu'/\underline{x}] \Rightarrow (\lambda = \lambda' \text{ and } \mu = \mu')$

$D_3$ (Weak independence)
$\pi_1 \succsim \pi_2 \Rightarrow \forall \alpha \in P_2, [\alpha_1/\pi_1, \alpha_2/\pi] \succsim [\alpha_1/\pi_2, \alpha_2/\pi]$.

$D_4$ (Continuity)
$\pi_1 \succ \pi_2 \succ \pi_3 \Rightarrow \exists \alpha \in P_2, [\alpha_1/\pi_1, \alpha_2/\pi_3] \sim \pi_2$.

Axiom $D_1$ is similar to axiom $C_1$. Axiom $D_2$ precludes the trivial case where all lotteries are equivalent. The other axioms are inspired by the set of axioms presented by Machina (1988). Axiom $D_3$ states that in a compound lottery replacing a sublottery by a preferred one cannot worsen that lottery. In sequential decision making, this axiom has a natural interpretation. It says that if an alternative is preferred to another one at one point in time, it will remain so at any other point. Remark that axiom $D_3$ stated with $k$ sublotteries would yield a stronger axiom than $C_3$ as it would imply it. However using explicitly this axiom allows this property of AEU to be highlighted. Axiom $D_4$ states that any two lotteries can be combined to be equivalent to any other lottery in between them. This axiom is stronger than $C_4$ and could be replaced by it. However using this stronger version underlines another property of AEU. These axioms are well-known in decision theory and focus on how the decision maker constructs his/her preferences. Note that in this set of axioms, the order over binary lotteries is not imposed. It will result from them.

For this set of axioms, we need to add two requirements to enable that some equations are solvable in semiring $P$.

$E_1$ $\forall \alpha, \beta \in P_2, \alpha \geq \beta \Rightarrow \exists \lambda \in P, \begin{cases} \alpha_1 = \beta_1 \oplus \lambda \\ \beta_2 = \alpha_2 \oplus \lambda \end{cases}$

$E_2$ $\forall \lambda, \mu \in P, \exists \alpha \in P_2, (\lambda \oplus \mu) \otimes \alpha = \langle \lambda, \mu \rangle$.

Condition $E_1$ says that for any two couples in $P_2$, it is possible to degrade the best one and to upgrade the worst one to get the same couple. Condition $E_2$ is a regularity requirement. Any sum of two elements of $P$ can be decomposed with a couple in $P_2$. These conditions entails that semiring $P$ behaves well. They are naturally satisfied for all the examples of plausibility measures given in Section 2, except for symbolic probability.

The two following lemmas will be useful to prove the second theorem. The first one states that under certain conditions, axiom $D_3$ implies $C_3$.

**Lemma 1** *Conditions $R$, $E_1$, $E_2$ and axiom $D_3$ $\Rightarrow$ $C_3$.*

**Proof.** First, we prove by recursion that condition $E_2$ can be stated on any number of elements: $\forall i = 1 \ldots k, \lambda_i \in P, \exists \alpha_i \in P, \bigoplus_{i=1}^{k} \alpha_i = 1_P$, s.t. $(\bigoplus_{i=1}^{k} \lambda_i) \otimes \alpha_i = \lambda_i, \forall i = 1 \ldots k$. This is true for $k = 2$ by $E_2$. Assume that it is true for $k$. Take $\forall i = 1 \ldots k+1, \lambda_i \in P$. Apply the case $k = 2$ on $\bigoplus_{i=1}^{k} \lambda_i$ and $\lambda_{k+1}$. We get $\beta \in P_2$ such that $(\bigoplus_{i=1}^{k+1} \lambda_i) \otimes \beta = \langle \bigoplus_{i=1}^{k} \lambda_i, \lambda_{k+1} \rangle$. Now apply the recursion assumption on $\lambda_i, \forall i = 1 \ldots k$. We get $\forall i = 1 \ldots k, \gamma_i \in P, \bigoplus_{i=1}^{k} \gamma_i = 1_P, \forall i = 1 \ldots k, (\bigoplus_{i=1}^{k} \lambda_i) \otimes \gamma_i = \lambda_i$. Then with $\forall i = 1 \ldots k, \alpha_i = \beta_1 \otimes \gamma_i$ and $\alpha_{k+1} = \beta_2$, the property is verified.

Now again by recursion over $k$, the lemma can be proved. For $k = 2$, this is true by axiom $D_3$. Assume that it is true for $k$. Take $\forall i = 1 \ldots k+1, \pi_i, \pi'_i \in \Pi^\infty(X), \pi_i \sim \pi'_i$ and $\forall i = 1 \ldots k+1, \lambda_i \in P, \bigoplus_{i=1}^{k} \lambda_i = 1_P$. Apply the extension of condition $D_2$ on $\lambda_i, \forall i = 1 \ldots k$. We get the $\alpha_i$. Apply the recursion assumption on the lotteries with indices $i = 1 \ldots k$ and the $\alpha_i$. We have $[\alpha_1/\pi_1, \ldots, \alpha_k/\pi_k] \sim [\alpha_1/\pi'_1, \ldots, \alpha_k/\pi'_k]$. Now apply the case $k = 2$ with $\bigoplus_{i=1}^{k} \lambda_i$ and $\lambda_{k+1}$. By axiom $R$, we obtain the desired property. ∎

The second lemma says that lotteries with a high plausibility on the best consequence and a low plausibility on the worst consequence are preferred. It gives one of the implications of axiom $C_2$.

**Lemma 2** *Under conditions $R$, $E_1$ and $E_2$, axioms $D_1$ and $D_3$, for any $\alpha, \beta \in P_2$, we have $(\alpha_1 \geq \beta_1 \text{ and } \beta_2 \geq \alpha_2)$ $\Rightarrow [\alpha_1/\overline{x}, \alpha_2/\underline{x}] \succsim [\beta_1/\overline{x}, \beta_2/\underline{x}]$.*

**Proof.** By assumption, $\overline{x} \succsim \underline{x}$. By $D_3$, for all $\pi$ in $\Pi(X)$, $(\lambda, \mu)$ in $P_2$, $[\lambda/\overline{x}, \mu/\pi] \succsim [\lambda/wX, \mu/\pi]$. Choose lottery $\pi = [\lambda'/\overline{x}, \mu'/\underline{x}]$. By condition $R$, $[\lambda \oplus \mu \otimes \lambda'/\overline{x}, \mu \otimes \mu'/\underline{x}] \succsim [\mu \otimes \lambda'/\overline{x}, \lambda \oplus \mu \otimes \mu'/\underline{x}]$. We want to solve the following system of equations:

$\begin{array}{llll} \alpha_1 = \lambda \oplus \beta_1 & (1) & \alpha_2 = \mu \otimes \mu' & (3) \\ \beta_1 = \mu \otimes \lambda' & (2) & \beta_2 = \lambda \oplus \alpha_2 & (4) \end{array}$

Equations 2 and 3 give $\mu = \alpha_2 \oplus \beta_1$. Conditions $E_1$ and $E_2$ give $\lambda, \lambda'$ and $\mu'$. ∎

The second representation theorem can then be stated.

**Theorem 2** *Under conditions $R$, $E_1$ and $E_2$, preference relation $\succsim$ over $\Pi^\infty(X)$ satisfies axioms $D_1$-$D_4$ iff there exists a basic utility assignment $u : X \to P_2$ such that $\forall \pi, \pi' \in \Pi^\infty(X), \pi \succsim \pi' \Leftrightarrow AEU(\pi) \geq_2 AEU(\pi')$.*

**Proof.** For the necessary part, assume that AEU represents the preference relation. Axioms $D_1$ and $D_2$ are naturally satisfied. Axiom $D_3$ results from the semiring structure of $P$ (distributivity of $\otimes$ over $\oplus$) and condition $R$. For axiom $D_4$, let $\pi_1 \succ \pi_2 \succ \pi_3$ with $AEU(\pi_1) = \alpha$, $AEU(\pi_2) = \beta$ and $AEU(\pi_3) = \gamma$. For $\langle \lambda, \mu \rangle \in P_2$, compute $AEU([\lambda/\pi_1, \mu/\pi_3]) = \langle \lambda \otimes \alpha_1 \oplus \mu \otimes \gamma_1, \lambda \otimes \alpha_2 \oplus \mu \otimes \gamma_2 \rangle$. We want to find a couple $\langle \lambda, \mu \rangle \in P_2$ such that $\lambda \otimes \alpha_1 \oplus \mu \otimes \gamma_1 = \beta_1$ and $\lambda \otimes \alpha_2 \oplus \mu \otimes \gamma_2 = \beta_2$ By assumption, $\alpha \geq_2 \beta \geq_2 \gamma$. Then there exist $\nu, \xi \in P$ such that

$\langle\alpha_1, \alpha_2\oplus\nu\rangle = \langle\beta_1\oplus\nu, \beta_2\rangle$ and $\langle\beta_1, \beta_2\oplus\xi\rangle = \langle\gamma_1\oplus\xi, \gamma_2\rangle$. Rearranging these equations, we get $\langle\alpha_1, \alpha_2\oplus\nu\oplus\xi\rangle = \langle\gamma_1\oplus\nu\oplus\xi, \gamma_2\rangle$. Using these equalities and assuming that $\lambda\oplus\mu = 1_P$, the system of equations becomes:
$$\begin{cases} \gamma_1\oplus\lambda\otimes(\nu\oplus\xi) = \gamma_1\oplus\xi \\ \alpha_2\oplus\mu\otimes(\nu\oplus\xi) = \alpha_2\oplus\nu \end{cases}$$
Then if a solution exists for these two equations:
$$\lambda\otimes(\nu\oplus\xi) = \xi \qquad \mu\otimes(\nu\oplus\xi) = \nu$$
a solution exists for the first system. By condition $E_2$, there exists $\langle\lambda, \mu\rangle \in P_2$ verifying the previous equations. This finishes the proof of axiom $D_4$.

Now we outline the proof of the sufficient part as it is quite classic. Axiom $D_4$ implies that for any degenerated lottery, i.e. any consequence, one can find an equivalent binary lottery. Then Lemma 1 and axiom $R$ entail that any lottery can also be reduced to a binary lottery. Axiom $D_1$ states that comparing lotteries is equivalent to comparing their associated binary lotteries. Finally axiom $D_2$ along with Lemma 2 gives the opposite implication of the lemma. This defines the order on $\Pi^\infty(X)$. ∎

Theorem 2 is based on simpler axioms than Theorem 1 but it needs two solvability conditions. Both theorems allow an AEU criterion to be derived as soon as one has an uncertainty representation satisfying our conditions. The obtained criterion would be the counterpart of EU in the chosen uncertainty representation. As an example of the interest of our result, we give two simple applications. Weydert (1994) proposed quasi-measures as general uncertainty measures. To the best of our knowledge, no decision model have been proposed for these measures. They satisfy all the conditions of Theorem 2. Then AEU can be constructed using quasi-measures. Blume et al. (1991) have considered lexicographic probabilities (which are vector of probabilities that are compared lexicographically) and derived lexicographic expected utility. It is easy to check that their model can be seen as an algebraic expected utility. In the same vein, one could consider lexicographic plausibility measures and derive lexicographic algebraic expected utility thanks to the previous theorem. Such decision models have not been studied yet.

## 4 PROPERTIES OF AEU

Algebraic expected utility satisfies the same properties as expected utility: linearity, dynamic consistency, autoduality of the underlying uncertainty representation, autoduality of the decision criterion and modeling of decision maker's attitude toward uncertainty[1].

---

[1] We thank Patrice Perny for the suggestion of this feature.

### 4.1 Linearity and Dynamic Consistency

First, obviously, AEU follows the linearity property. In the case of two consequences, this property states that for any two elements $\lambda, \mu \in P$ and any $x, x' \in X$, we have $AEU([\lambda/x, \mu/x']) = \lambda\otimes AEU(x)\oplus\mu\otimes AEU(x')$. This formula can of course be extended to any finite number of consequences. Remark that this is entailed by the independence axiom.

This property can be extended to $\Pi^\infty(X)$. For compound lotteries with two outcomes, it states that for any $\langle\lambda, \mu\rangle$ in $P_2$ and any two lotteries $\pi, \pi' \in \Pi^\infty(X)$, we have $AEU([\lambda/\pi, \mu/\pi']) = \lambda\otimes AEU(\pi)\oplus\mu\otimes AEU(\pi')$. Again, this can be easily extended to any compound lotteries with any number of outcomes. This property implies that AEU satisfies dynamic consistency. Dynamic consistency is a very nice property in sequential decision making. It states that if a decision maker prefers an alternative to another one at some step in the decision process then he/she will at any other step. This nice property allows AEU to be naturally integrated in sequential decision models, such as Markov decision processes (see Perny et al. (2005)).

### 4.2 Autoduality of the Uncertainty Representation

The *dual* of the preorder $\succsim_S$ over events $\mathcal{F}$ is the relation $\succsim_S^T$ defined by $\forall A, B \in \mathcal{F}, A\succsim_S^T B$ if and only if $\overline{B}\succsim_S\overline{A}$ where $\overline{A}$ and $\overline{B}$ denote the complements of $A$ and $B$. Naturally the dual of the dual of a preorder is the preorder itself.

Preorder $\succsim_S$ is said *autodual* if and only if for two events $A$ and $B$, $A\succsim_S B \Leftrightarrow \overline{B}\succsim_S\overline{A}$. For such autodual preorder, this means that $A$ is more plausible than $B$ if and only if the complement of $B$ is more plausible than that of $A$.

Duality and autoduality can be naturally extended to plausibility measures. A plausibility measure is the dual of another one if their induced preorders over events are duals of one another. As examples, possibility measures are the duals of necessity measures (Dubois and Prade (1980)) and belief and plausibility functions (Shafer (1976)) are the duals of one another. A plausibility measure is autodual if its induced preorder is autodual. A probability distribution is autodual.

An autodual measure gives more information than a non autodual one. Indeed if an autodual measure says that $A$ is more plausible than $B$ then it says also that the complement of $B$ is more plausible than that of $A$ and vice versa. In the case of a non autodual measure, there is no relation between the two facts: $A$ more plausible than $B$ and the complement of $B$ more plausible than that of $A$. And having $A$ more plausible than $B$ does not give any information on $\overline{A}$ and $\overline{B}$. Thus a non autodual plausibility measure does not use all the available information at hand since it could be refined with its dual measure. Therefore autodual plausibility measure should be exploited whenever possible.

Hopefully, from a non autodual measure Pl, there is an easy way to build an autodual measure. Indeed one just has to take the measure $\sigma_{Pl}$ defined by the couple $\sigma_{Pl}(A) = \langle Pl(A), Pl(\overline{A}) \rangle, \forall A \in \mathcal{F}$. These values are compared using order $\geq_2$. It can be checked that it is an autodual plausibility measure. In general, it does not give a complete order over events, even when Pl defines a complete order. For instance, this is the case with the couple of belief and plausibility functions. Note that if Pl is a probability distribution, then the measure defined by the couple $\sigma_{Pl}$ gives the same information as Pl and they are thus equivalent.

We now show that a decision maker using AEU has an autodual uncertainty representation. For two events $A, B$, consider binary acts $\overline{x}A\underline{x}$ and $\overline{x}B\underline{x}$ where act $\overline{x}A\underline{x}$ gives the best consequence $\overline{x}$ when event $A$ occurs and the worst consequence $\underline{x}$ otherwise and act $\overline{x}B\underline{x}$ gives $\overline{x}$ when $B$ occurs and $\underline{x}$ otherwise. Act $\overline{x}A\underline{x}$ is preferred to $\overline{x}B\underline{x}$ means that the decision maker believes that $A$ is more plausible than $B$. Thus comparing such binary acts defines a preorder over events. Note that $AEU(\overline{x}A\underline{x}) = \sigma_{Pl}(A) = \langle Pl(A), Pl(\overline{A}) \rangle$. Consequently even if the decision maker started with a non autodual plausibility measure Pl, he/she implicitly uses all the information provided by Pl to rank events when making decision with AEU with $\sigma_{Pl}$. This feature of AEU is particularly of interest when non autodual measures are used to represent uncertainty.

### 4.3 Autoduality of the Decision Criterion

The properties of duality and autoduality can be also stated on preference relation over acts. As the preorder over consequences is not assumed to be complete, the extension of the definitions is a bit tricky. We need to suppose that preferences over acts results from preferences over consequences. For any preorder $\succsim$ over acts $\mathcal{A}$ and any preorder $\succsim_X$ over consequences $X$, let $\succsim(\succsim_X)$ denote the preorder over acts when consequences are considered in their natural order and $\succsim(\precsim_X)$ the preorder over acts when consequences are considered in their reversed order, i.e. bad outcomes are supposed to be more attractive than good ones.

The *dual* of a preference relation $\succsim(\succsim_X)$, which is denoted $\succsim^T(\succsim_X)$ is defined by $f\succsim^T(\succsim_X)g \Leftrightarrow g\succsim(\precsim_X)f$. Roughly speaking when an act $f$ is preferred to an act $g$ in the sense of $\succsim^T(\succsim_X)$, it means that when attracted by bad consequences, $g$ is better ranked than $f$. Again, the dual of the dual of a preference relation is the relation itself. By extension, we say that a decision criterion is the dual of another one if the preference relation defined by the former decision criterion is the dual of the relation defined by the latter. As an example, pessimistic utility is the dual of optimistic utility.

A preference relation is *autodual* if and only if $\succsim^T(\succsim_X) = \succsim(\succsim_X)$, which means that for any two acts $f, g$, $f\succsim(\succsim_X)g \Leftrightarrow g\succsim(\precsim_X)f$. Again, by extension, a decision criterion is autodual if its induced preference relation is autodual. Intuitively, this means that the decision criterion when comparing two acts looks at how well and how bad these two acts perform. As previously, any non autodual decision criterion does not use all the information at hand since it could be possible to refine with the dual of the decision criterion. In this sense a rational decision maker would rather use an autodual decision criterion. As an example, EU is autodual. Being defined as a couple, one element focusing on the best consequence and the other on the worst one, it can be easily checked that AEU is also autodual.

### 4.4 Modeling of Decision Maker's Attitude

With EU, attitude towards risk can be integrated to the model by means of the shape of the basic utility assignment. When the utility assignment is concave, the decision maker is risk averse. He/she would prefer sure consequences to risky consequences. On the contrary when the utility assignment is convex, the decision maker is risk-seeking. The modeling of the decision maker attitude toward risk relies on how utility is affected to consequences. By shaping the basic utility assignment, one can integrate the decision maker's attitude into the decision model.

This idea can be naturally transposed in AEU. The choice of the basic utility assignment allows the decision maker's attitude to be taken into account. We illustrate this property by a simple example. Let $y_1, y_2, y_3$ three consequences such that $y_1\succ_X y_2\succ_X y_3$. Let $\pi = [\lambda/y_1, \mu/y_3]$. Let $u^= : X \rightarrow P_2$ a basic utility assignment such that $AEU_{u^=}(\pi) = u(y_2)$. A decision maker whose preferences and attitude toward uncertainty are modeled by $u^=$ considers that uncertain alternative $\pi$ and sure consequence $y$ are equivalent. Now consider $u^+ : X \rightarrow P_2$ a basic utility assignment such that $\forall x \in X, u^+(x) = u^=(x)$ if $y_2\succsim_X x$ and $u^+(x)\geq_2 u^=(x)$ otherwise. Then obviously, $AEU_{u^+}(\pi)\geq_2 u^+(y_2)$ by monotony of the operators. A decision maker whose preferences are modeled by $u^+$ will favor the uncertain alternative to the sure consequence. In this sense, he/she is more uncertainty-seeking than a decision maker using $u^=$. On the contrary, with $u^- : X \rightarrow P_2$ a basic utility assignment such that $\forall x \in X, u^-(x) = u^=(x)$ if $x\succsim_X y_2$ and $u^=(x)\geq_2 u^-(x)$ otherwise, sure consequence $y_2$ will be preferred. Such a attitude is more uncertainty averse than that obtained with $u^=$.

This is exactly the case with binary possibilistic utility. Giang and Shenoy (2001) showed that when all the basic utilities are high, binary possibilistic utility reduces to pessimistic utility. On the contrary, when all the basic utilities are low, binary possibilistic utility reduces to optimistic utility. In addition to their result, our work suggests that the degree of optimism or pessimism can be controlled more finely by how much the basic utilities are shifted.

## 5 RELATED WORKS

Many proposed decision models are instances of algebraic expected utilities. We list here some examples:

- standard expected utility,
- lexicographic expected utility introduced by Blume et al. (1991),
- the decision criteria proposed by Giang and Shenoy (2000, 2001, 2002) for respectively $\kappa$-rankings, qualitative possibility and quantitative possibility,
- optimistic and pessimistic utilities (Dubois and Prade (1995)) as they are particular cases of the criterion proposed by Giang and Shenoy (2001),
- maximax and maximin criteria (axiomatized by Brafman and Tennenholtz (1997)) as they are particular cases of optimistic and pessimistic utilities,
- pessimistic utility refined by optimistic utility axiomatized by Dubois et al. (2000a) as proved by Weng (2005), it is a binary utility,
- refined possibilistic utility proposed by Weng (2005),
- the decision model proposed by Wilson (1995) on extended reals. Our work gives an axiomatic foundation of his proposal.
- the decision criterion proposed by Giang and Sandilya (2004) as symbolic probability is a plausibility measure.

Giang and Shenoy (2003) proposed a decision criterion for partially consonant belief functions, which are generalizations of both probability and possibility measures. More specifically, they considered partially consonant belief functions in the form of probability distributions over possibility distributions over consequences. For these two-level measures, they applied the idea of binary utility twice getting a decision criterion that is formally close to algebraic expected utility, as it can be seen as a two-level AEU. This suggests that we could define more general decision criteria using plausibility distribution defined over plausibility distributions over consequences and applying Theorem 2 twice. An interesting work would be to study the possibility to find a one-level plausibility measure that would yield an AEU equivalent to a two-level AEU.

Giang and Sandilya (2004) recently proposed a general decision model similar to ours using symbolic probability as uncertainty representation. They lead their axiomatization in a setting similar to Schmeidler (1989) who used the framework presented by Anscombe and Aumann (1963). This setting is justified if one wants to justify axiomatically the use of subjective probability or more generally the use of subjective plausibility measures. Indeed in the context of decision problems, Anscombe and Aumann derived subjective additive probabilities and axiomatized subjective expected utility. Schmeidler generalized the result by relaxing some of these axioms and obtained subjective non-additive probabilities and Choquet expected utility (CEU). However in this setting, Giang and Sandilya (2004) did not derive subjective symbolic probabilities. Thus the choice of this setting is not, in our opinion, the most suitable for an axiomatization of a general criterion in contexts where plausibility measures are given. In this case, a vNM setting is a preferable framework. Consequently, our two axiomatizations are formulated in such a setting. The first is very similar to theirs as we simply restated their axioms in a vNM setting. The second axiomatization requiring two solvability requirements is based on simpler axioms and give a better understanding of the properties of AEU. In this second set of axioms, $B_2$, which enforces the order on binary lotteries, is not artificially imposed. This order naturally results from the set of axioms, i.e. from the behavior of the decision maker.

When the scale is $[0, 1]$, our semiring requirements implies that operators $\oplus$ and $\otimes$ are respectively a t-conorm and a t-norm. Decomposable plausibility measures in such contexts and integrals of the form of AEU have been extensively studied (Klement et al. (2000); Dubois et al. (1996)). Interestingly, in this totally ordered case, the semiring requirements, especially that of distributivity, enforce $+/\times$, $\max$/t-norm mixtures or ordinal sums of these two (Dubois et al. (2000c,b)). AEU encompasses these integrals and our work gives an axiomatic foundations (in a decision-theoretic sense) to them as a decision criterion. Our proposal is stated on more general scales, which can be only partially ordered. It would be interesting to investigate the possibility of revealing the family of mixtures enforced by the semiring requirements in this case.

As a final remark, our work shows that AEU is formally different of CEU as CEU uses a non additive plausibility measure. Therefore there is no operator $\oplus$ and $\otimes$ satisfying the required conditions for AEU.

## 6 CONCLUSION

We provided two axiomatic justifications for a class of generalized expected utility that we named algebraic expected utility. These axiomatizations have been lead in a vNM setting, i.e. the uncertainty representation is supposed to be given and it is here assumed to be captured by a plausibility measure. Our second axiomatization is particularly of interest. Though requiring two solvability conditions, it is based on simple and natural axioms (preorder, non triviality, weak independence and continuity), which are similar to those presented in Machina (1988) for EU.

Our general algebraic approach allowed all the previous propositions similar to expected utility to be unified in a same formalization. From any uncertainty representation satisfying our conditions, an algebraic expected utility can be constructed. This is especially of interest for uncertainty representation for which decision models have not yet been proposed.

We showed that having standard expected utility as a special case, AEU enjoys the same nice properties as EU: linearity, dynamic consistency, autoduality of the underlying uncertainty representation, autoduality of the decision criterion, modeling of decision maker's attitude towards uncertainty. As a side point, it can be highlighted that this decision criterion can be used in sequential decision making, especially with Markov decision processes (see Perny et al. (2005)).

As a future work, we plan to generalize AEU in the same direction as Schmeidler did with EU to get CEU. Many experiments showed that axioms of EU are violated in many situations. This proves therefore that EU cannot be used as a criterion to model these decision situations. In these cases, CEU can be used as it has a much higher power of description than EU. Generalization of AEU in the direction of CEU should also offer a greater flexibility in decision behavior modeling. Interestingly, in the case of possibilistic utility, optimistic and pessimistic utilities can be generalized to get Sugeno integrals, which are qualitative counterpart of CEU.